\pdfoutput=1
\documentclass[11pt,svgcolors]{article}

\usepackage[preprint]{acl}

\usepackage{times}
\usepackage{latexsym}
\usepackage{multirow}

\usepackage[T1]{fontenc}

\usepackage[utf8]{inputenc}

\usepackage{colortbl}

\usepackage{microtype}

\usepackage{inconsolata}

\usepackage{graphicx}

\usepackage{tipa}

\newcommand\Language[2]{\href{https://glottolog.org/resource/languoid/id/#1}{#2}}
\newcommand\mycite[1]{{\ttfamily #1} \citep{#1}}

\title{Unstable Grounds for Beautiful Trees? Testing the Robustness of Concept Translations in the Compilation of Multilingual Wordlists}

\author{
  \textbf{David Snee\textsuperscript{1}},
  \textbf{Luca Ciucci\textsuperscript{1}},
  \textbf{Arne Rubehn\textsuperscript{1}},
  \textbf{Kellen Parker van Dam\textsuperscript{1}}, \\
  \textbf{Johann-Mattis List\textsuperscript{1}}
\\
\\
  \textsuperscript{1}Chair for Multilingual Computational Linguistics, University of Passau, Passau, Germany \\
\\
  }

\begin{document}
\maketitle
\begin{abstract}
Multilingual wordlists play a crucial role in comparative linguistics. While many studies have been carried out to test the power of computational methods for language subgrouping or divergence time estimation, few studies have put the data upon which these studies are based to a rigorous test. Here, we conduct a first experiment that tests the robustness of concept translation as an integral part of the compilation of multilingual wordlists. Investigating the variation in concept translations in independently compiled wordlists from 10 dataset pairs covering 9 different language families, we find that on average, only 83\% of all translations yield the same word form, while identical forms in terms of phonetic transcriptions can only be found in 23\% of all cases. Our findings can prove important when trying to assess the uncertainty of phylogenetic studies and the conclusions derived from them. 
\end{abstract}

\section{Introduction}


While the quantitative turn in historical linguistics has been met with a considerable amount of skepticism for a long time \citep{Holm2007,Geisler2022}, phylogenetic methods -- originally developed to infer phylogenies of biological species -- have by now become the new state of the art in the field, replacing more traditional methods for subgrouping almost completely. Given that most linguistic approaches to phylogenetic reconstruction make use of lexical data, \emph{multilingual wordlists} play a crucial role in phylogenetic reconstruction in historical linguistics. 

The compilation of multilingual wordlists itself is quite tedious. Starting from a list of concepts, scholars must translate the concepts into all target languages under investigation. The translation into the target languages, however, is no standardized procedure, but may require various steps, including the consultation of informants, the consultation of published resources, or the inspection of archived material. In all these cases, scholars who compile a wordlist must weight their evidence carefully, in order to avoid errors. Given the complexity of this process, it is no surprise that errors can slip in easily into the translations. A given concept may lack a direct translational equivalent in a given language, or there may be several good candidates from which scholars must select the most appropriate ones. As a result, there is a great risk that multilingual wordlists compiled for phylogenetic studies show a considerable amount of idiosyncrasies that might have an impact on the phylogenies scholars compute from them.

Studies that try to measure the amount of inconsistency in multilingual wordlists -- introduced by the translation of concepts into target languages -- are lacking so far. Here, we present a first attempt to shed light on the robustness of the concept translation task, taking advantage of the fact that recent efforts have produced large-scale repositories of standardized multilingual wordlists \citep{List2022e}. In the following, we will give a short overview on previous discussions and studies that focus on the translation of concepts in multilingual wordlist compilation (§~\ref{sec:background}). After this, we introduce the materials and methods by which we try to evaluate the robustness of concept translation (§~\ref{sec:mm}). Having presented the results (§~\ref{sec:res}), we discuss them in more detail and share some ideas to improve the enterprise of wordlist compilation in historical linguistics (§~\ref{sec:con}).




\section{Background}\label{sec:background}


Phylogenetic approaches rely on multilingual wordlists compiled through lexicostatistic methods, dating back to Swadesh's foundational work \citep{Swadesh1950, Swadesh1952, Swadesh1955}. A multilingual wordlist in this context is a list of concepts translated into one or more target languages \citep{List2014d}. 
Although typically emphasizing their difference with respect to Swadesh's lexicostatistics, modern phylogenetic approaches all build on this onomasiological (i.e. \emph{concept-based}) approach that takes the concept as the major aspect by which languages are compared.
Building on \citet{Geisler2010},  we can identify five major steps in the typical workflow applied in modern approaches to phylogenetic reconstruction. Starting from the compilation of a concept list~(1: \emph{concept list compilation}), the concepts are translated into the target languages~(2: \emph{concept translation}) in order to create an initial \emph{comparative wordlist}. This wordlist is then used as the basis for the identification of cognate words~(3: \emph{cognate identification}). Having converted the information on cognate words into a numerical or computer-readable format~(4: \emph{cognate coding}), scholars then employ their phylogenetic method of choice in order to compute a phylogeny of the languages in question~(5: \emph{phylogenetic reconstruction}). 

\begin{table*}[tb]
\centering
\resizebox{\textwidth}{!}{
\begin{tabular}{|l|p{4cm}|p{4cm}|rrr|rrr|rr|}
\hline
 \multirow{2}{*}{Group}        & \multirow{2}{*}{Dataset A} & \multirow{2}{*}{Dataset B}              &   \multicolumn{3}{c|}{Concepts}&   \multicolumn{3}{c|}{Languages} &   \multicolumn{2}{|c|}{Synonymy} \\
 & & & A & B & A{$\cap$}B & A & B & A{$\cap$}B & A & B\\ 
\hline
 Bai           & \mycite{allenbai} \textcolor{white}{.....}    & \mycite{wangbai}               &         499 &         412 &              208 &            9 &           10 &                 2 &     1.01 &     1.02 \\\hline
 Chadic        & \mycite{kraftchadic}  & \mycite{gravinachadic}         &         429 &         717 &              325 &           67 &           48 &                 5 &     1.04 &     1.11 \\\hline
 Chinese       & \mycite{beidasinitic} & \mycite{liusinitic}            &         738 &         202 &              102 &           18 &           19 &                12 &     1.18  &     1.17 \\\hline
 Dravidian     & \mycite{dravlex}      & \mycite{northeuralex}          &         100 &         954 &               94 &           20 &          107 &                 4 &     1.39 &     1.01 \\\hline
 Indo-European & \mycite{iecor}        & \mycite{starostinpie}          &         170 &         110 &               88 &          160 &           19 &                15 &     1.01 &     1.04 \\\hline
 Japonic       & \mycite{leejaponic}   & \mycite{robbeetstriangulation} &         210 &         254 &              152 &           59 &          101 &                 6 &     1.01 &     1.02 \\\hline
 Koreanic      & \mycite{leekoreanic}  & \mycite{robbeetstriangulation} &         246 &         254 &              175 &           15 &          101 &                13 &     1.01  &     1.01 \\\hline
 Tupian        & \mycite{galuciotupi}  & \mycite{gerarditupi}           &         100 &         242 &               70 &           23 &           38 &                 5 &     1.02 &     1.00       \\\hline
 Uralic        & \mycite{northeuralex} & \mycite{syrjaenenuralic}       &         954 &         173 &              147 &          107 &            7 &                 5 &     1.12 &     1.17 \\\hline
 Uto-Aztecan   & \mycite{utoaztecan}   & \mycite{davletshinaztecan}     &         121 &         100 &               92 &           46 &            9 &                 3 &     1.22 &     1.00 \\
\hline
\end{tabular}}
\caption{Selected language groups along with their original datasets and additional statistics employed in this study. References to the datasets follow the most recent publication of the data as part of the Lexibank repository. Information on the original studies in which the data were published for the first time are provided by the more recent standardized editions. The table lists the number of concepts and languages (along with the intersection), as well as the synonymy (measured by dividing the number of words by the number of concepts).}
\label{tab:data}
\end{table*}

Up to now, most critics of lexicostatistics and its modern equivalents have concentrated on either the stage of cognate identification or the resulting phylogenetic methods. Cognate identification is often criticized as being flawed due to undetected borrowings \citep{Donohue2012}. When disputing over phylogenetic methods, there have been long-standing debates about the complexity of the models employed, as reflected in the debate about the age of Indo-European, where models differing in complexity yielded quite different age estimates \citep{Bouckaert2012,Chang2015,Kassian2021b,Heggarty2023}.

What has much less often been discussed in the context of phylogenetic methods, however, are the first two stages of the workflow, that is, the stage of concept list compilation, and the stage of concept translation. While it has been clear for a long time that different concept lists often yield different phylogenies \citep{Chen1996, McMahon2005b}, a closer discussion regarding the impact of concept lists on the results of phylogenetic analyses has not been carried out so far. The same holds for the translation of concepts into target languages. While \citet{Geisler2010} found that concept translation across Romance languages in two independently compiled multilingual wordlists 
differs by about 10\%, and \citet{List2018PBLOG2} and \citet{Haeuser2024} could show that the reduction of multiple translations for the same concept can have direct consequences on the resulting phylogenies, no closer investigation regarding the degree of variation in concept translation or the impact of concept translation on phylogenetic analyses has been conducted up to now.

\begin{figure*}[tb]
\centering
\includegraphics[width=\textwidth]{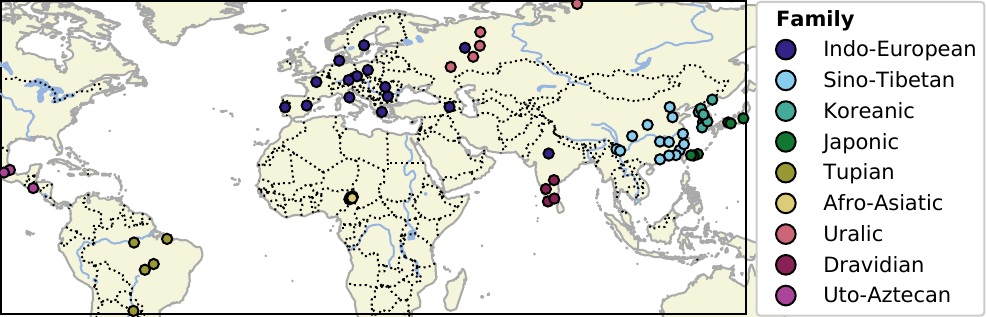}
\caption{Location of the languages investigated in our study. For each of the 70 languages, two wordlists were identified in the Lexibank repository.}
\label{fig:map}
\end{figure*}
 
\section{Materials and Methods}\label{sec:mm}

\subsection{Materials}
In order to investigate variation in word choice across multilingual wordlists, it is important to find wordlists that have been compiled independently for the same language varieties, containing at least a certain subset of identical concepts. In order to identify such data, we checked datasets published as part of the Lexibank repository (\href{https://lexibank.clld.org}{https://lexibank.clld.org}, \citealt{List2022e}), searching specifically for those cases where several languages from the same language family or subgroup are available in the form of multilingual wordlists created by different authors. 
 
Lexibank uses Cross-Linguistic Data Formats (CLDF, \href{https://cldf.clld.org}{https://cldf.clld.org}, \citealt{Forkel2018}) to standardize multilingual wordlists along the three dimensions of language, meaning, and form. Languages are linked to Glottolog (\href{https://glottolog.org}{https://glottolog.org}, \citealt{Glottolog}) in order to ensure that languages can be easily identified across sources, even if they are given different names in the original datasets. Concepts are mapped to Concepticon, a reference catalog for semantic glosses used to elicit concepts in concept lists (\href{https://concepticon.clld.org}{https://concepticon.clld.org}, \citealt{Concepticon}), in order to facilitate the aggregation of wordlists from different sources to allow the identification of common concepts for which different wordlists provide translations in their target languages. Phonetic transcriptions in Lexibank are unified with the help of the Cross-Linguistic Transcription Systems reference catalogue (CLTS, \href{https://clts.clld.org}{https://clts.clld.org}, \citealt{CLTS}), a standardized subset of the International Phonetic Alphabet that has a generative component by which detailed transcriptions of more than 8000 speech sounds can be created and compared (see \citealt{Anderson2018} and \citealt{Rubehn2024}).

Checking the Lexibank data in the most recent version of the repository (2.0, \citealt{Blum2025}), we identified 10 groups of languages corresponding to 9 different languages families, in which two and more multilingual wordlists from different datasets could be compared. The ten groups stem from 18 different datasets, with two datasets (\texttt{NorthEuralex}, see \citealt{Dellert2020}, and \texttt{RobbeetsTriangulation}, see \citealt{Robbeets2021}) offering two groups each. 
From these 10 wordlist pairs, each consisting of two multilingual wordlists covering at least two language varieties, we manually selected 70 language pairs, making sure that all pairs represent identical languages to the best of our knowledge. Table \ref{tab:data} provides an overview on the ten groups of language pairs that we compiled for this study, along with the number of matching concepts, matching language pairs, and synonymy statistics on the wordlists. Figure \ref{fig:map} shows the distribution geographical distribution of the 70 languages in our sample.


\subsection{Wordlist Comparison}


We compare wordlists in terms of their (1)~matching Glottocodes, (2)~manually selected language pairs, and (3)~matching concepts. Paired language varieties within each wordlist comparison are initially identified based on matching Glottocodes. This results in a total of 75 Glottocode matches across all language families to be compared. Upon closer examination of the data, it becomes clear that Glottocode comparisons alone do not allow for the comparison of identical language varieties as different subvarieties are at times encoded with the same Glottocode both inside the same dataset and in separate datasets. To quantify the effect of this discrepancy, language pairs are also manually selected based on consultation with each dataset's metadata, resulting in a total of 70 language pairs for analysis. Language pairs are then examined based on matching concepts, which are identified with the help of the Concepticon mappings provided by Lexibank.

\subsection{Comparing Concept Translations}

Since phonetic transcriptions in Lexibank's datasets are unified, following the system recommended by the CLTS reference catalog, one might expect that differences in concept translation can be simply identified by comparing transcriptions across different datasets directly, using string identity as criterion to assess if two translations are identical or not. However, word form comparisons using phonetic data are still error-prone as datasets often differ with respect to details in the concrete realization of phonetic transcriptions \citep{Anderson2018}. Since variation in phonetic transcription is rather a norm than an exception \citep{Anderson2023}, we have to find a metric that allows us to distinguish those cases where two translations are identical even if the phonetic transcriptions differ slightly from those cases where two translations point to two different words.
In order to address this problem, we decided for an automated approach that measures the phonetic similarity of sound sequences, rather than their identity with respect to the symbols used in transcriptions.

Our approach makes use of the Sound-Class Based Phonetic Alignments algorithm \citep[SCA,][]{List2012c}, which was originally designed to align words phonetically, but which comes along with a measure for phonetic distances that ignores minor transcription differences. SCA distance scores derived from phonetic alignments carried out with the help of the SCA algorithm, have been shown to work quite well in the task of automated cognate detection \citep{List2012b} and borrowing identification \citep{Miller2023}. SCA distance scores range between 0 (near identity of phonetic sequences) and 1 (very low similarity). Assuming that a score below 0.5 points to differences resulting from phonetic transcriptions, while scores higher than 0.5 result from differences stemming from different translations, we employ SCA distances as a proxy to detect whether two translations are \emph{similar} (reflecting only phonetic variation) or \emph{different} (reflecting true differences in translation). 
 
In addition to the automated identification of translation differences with the help of SCA distances, we also computed the average SCA distances for all language pairs in our sample, as well as the edit distance \citep[also known as \emph{Levenshtein distance)},][]{Levenshtein1966}, both in its original and its normalized form (where we divide the distance by the longer of the two sequences, see \citealt[178]{List2014d}). All in all these metrics should allow us to assess the amount of differences in concept translations across multilingual wordlists fairly well.

\subsection{Preprocessing}
Lexibank occasionally contains data in which morpheme boundaries are marked with the help of a plus symbol (\texttt{+}). Since morpheme boundaries would unnecessarily confuse phonetic alignment algorithms, introducing extra noise that we are not primarily interested in, we deleted all morpheme boundary markers from the sound sequences before comparing them. For the same reason, we also decided to ignore tone markers in the data. These do not occur in all datasets, but are instead mostly restricted to South-East Asian languages. However, since tone annotation in phonetic transcription can vary considerably, probably even more than the transcription for consonants and vowels, we also ignored all tones in datasets from South-East Asian languages. 

\subsection{Evaluation}
In order to test whether this approach is suitable to provide useful information on translation differences, we created a test set. Using EDICTOR \citep{EDICTOR}, translation differences were annotated for all language and concept pairs in the Indo-European datasets of our sample. This allows us to use the manual annotations as a gold standard and to assess the suitability of SCA distance scores to identify true differences in concept translation. Results of this comparison are reported in the form of precision, recall, and F-Scores \citep[191--192]{List2014d}.

\subsection{Implementation}
All methods are implemented in Python. We use LingPy to compute SCA distances and edit distances \citep[\href{https://pypi.org/project/lingpy}{https://pypi.org/project/lingpy}, Version 2.6.13,][]{LingPy}. For the handling of wordlist data in CLDF format, we use CL Toolkit \citep[\href{https://pypi.org/project/cltoolkit}{https://pypi.org/project/cltoolkit}, Version 0.2.0,][]{CLToolkit}. For the inspection and manual annotation of wordlist data, EDICTOR was used \citep{EDICTOR}. For the creation of the map in Figure \ref{fig:map}, CLDFViz was used \citep[Version 1.3.0,][]{CLDFViz}.

\section{Results}\label{sec:res}

\subsection{Handling Phonetic Variation}
As mentioned before,
we are interested in finding a way to identify words that reflect the same original word form that is represented in slight variations in the phonetic transcriptions. Having a reliable automated approach for this task is important in order to allow us to investigate differences in concept translation on a larger amount of data.
 
To test, how well the SCA distances can help us to identify words that only vary by phonetic transcription, we therefore conducted a test study in which we annotated all hand-selected language pairs from the Indo-European group, indicating for each pair of words whether they were identical -- despite potentially diverging phonetic transcriptions -- or different. This dataset of 15 manually annotated language pairs was then compared with the results derived from our automated approach using the SCA distance. The results of this analysis are shown in Table \ref{tab:ie}. As can be seen from the table, there are only minor differences between the automated and the manual annotation, with F-Scores of 0.98 on average for all 15 language pairs. We conclude from this that using SCA distances with a threshold of 0.5 provides a very good approximation to distinguish between identical word forms with potentially diverging transcriptions and word forms that reflect different word forms, allowing us to derive major conclusions when testing it on additional datasets.

\begin{table}[t]
\resizebox{\linewidth}{!}{
\centering
\begin{tabular}{|lrrr|}
\hline
Language  &   Precision &   Recall &   F-Score \\
\hline
 \Language{nucl1235}{Armenian (Eastern)}     &        0.99 &     0.99 &      0.99 \\
 \Language{bulg1262}{Bulgarian}     &        1.00 &     0.99 &      0.99 \\ 
 \Language{czec1258}{Czech}     &        1.00 &     0.99 &      1.00 \\
 \Language{dani1285}{Danish}     &        1.00 &     0.95 &      0.97 \\
 \Language{stan1290}{French}     &        0.99 &     0.99 &      0.99 \\
 \Language{stan1295}{German}     &        1.00 &     0.95 &      0.98 \\
 \Language{mode1248}{Greek}     &        0.97 &     0.99 &      0.98 \\
 \Language{hind1269}{Hindi}    &        0.99 &     0.99 &      0.99 \\
 \Language{poli1260}{Polish}     &        1.00 &     0.99 &      1.00 \\
 \Language{port1283}{Portuguese}     &        1.00 &     0.96 &      0.98 \\
 \Language{roma1327}{Romanian}     &        0.98 &     0.96 &      0.97 \\
 \Language{russ1263}{Russian}     &        1.00 &     0.99 &      0.99 \\
 \Language{stan1288}{Spanish}     &        1.00 &     0.98 &      0.99 \\
 \Language{swed1254}{Swedish}     &        0.99 &     0.97 &      0.98 \\
 \Language{ital1282}{Italian}     &        0.97 &     0.96 &      0.96 \\
 \hline
 TOTAL        &        0.99 &     0.98 &      0.98 \\
\hline
\end{tabular}}
\caption{Comparison of the evaluation study on Indo-European. Precision and recall are calculated per concept, counting true and false negatives and positives for all possible pairings within the same concept slot for identical languages. F-Scores are based on the harmonic mean calculated from precision and recall.}
\label{tab:ie}

\end{table}

\begin{table*}[tb]
\centering
\resizebox{\textwidth}{!}{%
\begin{tabular}{|lr|rr|rr|rr|rr|rr|}

\hline
 Family        &   Pairs &  $\uparrow$ Identical &   STD &  $\uparrow$ Similar &   STD &  $\downarrow$ SCA &   STD &  $\downarrow$ NED &   STD &  $\downarrow$ ED &   STD \\
\hline
 Bai           &       2 &        0.32 &     0.12 &      0.83 &      0.02 &  0.20 &      0.04 &  0.44 &      0.10 & \bfseries 1.33 &     0.32 \\
 Chadic        &      10 &        0.07 &     0.05 &      \cellcolor{lightgray} 0.71 &      0.14 &  \cellcolor{lightgray} 0.32 &      0.10 &  0.54 &      0.11 & \cellcolor{lightgray} 3.34 &     0.63 \\
 Chinese       &      12 &        \bfseries 0.39 &     0.11 &      0.77 &      0.05 &  0.24 &      0.04 &  0.38 &      0.06 & 1.66 &     0.25 \\
 Dravidian     &       4 &        0.06 &     0.04 &      0.79 &      0.02 &  0.25 &      0.03 &  0.57 &      0.08 & 3.30 &     0.44 \\
 Indo-European &      15 &        0.31 &     0.22 &      \bfseries 0.94 &      0.03 &  \bfseries 0.09 &      0.05 &  0.30 &      0.13 & 1.48 &     0.56 \\
 Japonic       &       7 &        0.28 &     0.19 &      0.74 &      0.24 &  0.27 &      0.17 &  0.40 &      0.20 & 2.09 &     1.02 \\
 Koreanic      &       9 &        \cellcolor{lightgray} 0.06 &     0.04 &      0.74 &      0.21 &  0.31 &      0.14 &  \cellcolor{lightgray} 0.59 &      0.12 & 3.01 &     0.72 \\
 Tupian        &       6 &        0.32 &     0.19 &      0.73 &      0.32 &  0.29 &      0.21 &  0.39 &      0.25 & 2.01 &     1.22 \\
 Uralic        &       6 &        0.19 &     0.12 &      0.85 &      0.06 &  0.17 &      0.08 &  0.42 &      0.13 & 2.26 &     0.78 \\
 Uto-Aztecan   &       4 &        0.34 &     0.22 &      0.88 &      0.03 &  0.18 &      0.06 &  \bfseries 0.32 &      0.11 & 1.79 &     0.85 \\
 TOTAL         &      75 &        0.23 &     0.13 &      0.80 &      0.07 &  0.23 &      0.07 &  0.44 &      0.10 & 2.23 &     0.74 \\
\hline
\end{tabular}}
\caption{Major results per dataset for our comparative study, comparing all languages that show matching Glottocodes with each other, in terms of Identical (phonetic strings match perfectly) and Similar word pairs (phonetic strings show SCA distance beyond our threshold of 0.5), as well as averaged SCA distances, normalized edit distance, and traditional edit distances. Highest similarities are marked in bold font, lowest similarities are shaded in gray. }
\label{tab:comp}
\end{table*}

\subsection{Variation in Concept Translation}
We conducted two tests on the ten datasets. During the first test, we compared all languages by matching Glottocodes with each other. If more than one variety was assigned the same Glottocode in a given dataset, all possible pairs were assembled and average values for word pair identity, similarity, and SCA distances were computed. In the second experiment, only those language pairs were considered that we had identified as reflecting the same varieties (to the best of our knowledge). In all cases, we computed the identity of the sound sequences that appeared as translations for the same concepts, the \emph{similarity} (with those pairs defined as similar whose SCA distance was below our threshold of 0.5), and the SCA distances. If more than one translation was available for a given concept in a given language variety, all possible pairs were compared and the average value of the individual scores was computed. 

The results of this comparison are shown in Table \ref{tab:comp} for the comparison based on matching Glottocodes and in Table \ref{tab:comp2} for the comparison based on hand-selected language pairs. As can be seen from the tables, the results do not differ too much from each other, at least as far as their tendencies are concerned. 
Nevertheless, a closer inspection of the differences between the two approaches shows that the linking of languages to Glottocodes shows some major problems in the datasets on the Chadic, Japonic, and Koreanic groups. In all three dataset pairs (Japonic and Koreanic data in one pair are both taken from the same study of \citealt{Robbeets2021}), we find that links to Glottolog could be improved. The data by \citet{Robbeets2021}, for example, provides the same Glottocodes for historical and modern varieties of Japanese and Korean. Any comparison that matches solely by Glottocode will therefore run the risk of comparing data from different stages of the same language. This example shows that scholars who base their analyses on Glottocodes should take particular care in selecting the most representative varieties. Especially when aggregating languages from different sources, one should make sure to provide extra checks on top of CLDF that would make sure that the same language varieties are being compared. Matching Glottocodes are an extremely good proxy, but they do not provide a guarantee that languages varieties from different sources really match. 

What we can also see from the table is that we have extremely low values on phonetic identity in both comparisons, while phonetic similarity (as reflected in SCA distance scores beyond the value 0.5) shows a drastic increase. This proves the usefulness of computing SCA distance scores instead of comparing if sound sequences are fully identical or not. It also shows (as we have already seen in the previous section) that SCA distances seem to provide quite sensitive results when it comes to assessing the near-identity of sound sequences reflecting slight transcription differences. 

\begin{table*}[tb]
\centering
\resizebox{\textwidth}{!}{%
\begin{tabular}{|lr|rr|rr|rr|rr|rr|}
\hline
 Family        &   Pairs &   $\uparrow$ Identical &   STD &  $\uparrow$ Similar &   STD &  $\downarrow$ SCA &   STD &  $\downarrow$ NED &   STD &  $\downarrow$ ED &   STD \\
\hline
 Bai           &       2 &        0.32 &     0.12 &      0.83 &      0.02 &  0.20 &      0.04 &  0.44 &      0.10 & \bfseries 1.33 &     0.32 \\
 Chadic        &       5 &        0.11 &     0.03 &      0.79 &      0.04 &  \cellcolor{lightgray} 0.27 &      0.02 &  \cellcolor{lightgray} 0.48 &      0.06 & \cellcolor{lightgray} 3.00 &     0.36 \\
 Chinese       &      12 &        \bfseries 0.39 &     0.11 &      \cellcolor{lightgray} 0.77 &      0.05 &  0.24 &      0.04 &  0.38 &      0.06 & 1.66 &     0.25 \\
 Dravidian     &       4 &        0.06 &     0.04 &      0.79 &      0.02 &  0.25 &      0.03 &  0.57 &      0.08 & 3.30 &     0.44 \\
 Indo-European &      15 &        0.31 &     0.22 &      \bfseries 0.94 &      0.03 &  \bfseries 0.08 &      0.03 &  \bfseries 0.29 &      0.11 & 1.39 &     0.51 \\
 Japonic       &       6 &        0.31 &     0.18 &      0.83 &      0.05 &  0.21 &      0.07 &  0.34 &      0.10 & 1.75 &     0.54 \\
 Koreanic      &      13 &        \cellcolor{lightgray} 0.06 &     0.03 &      0.82 &      0.09 &  0.25 &      0.06 &  0.54 &      0.06 & 2.76 &     0.45 \\
 Tupian        &       5 &        0.38 &     0.13 &      0.86 &      0.05 &  0.21 &      0.05 &  \cellcolor{lightgray} 0.29 &      0.05 & 1.53 &     0.34 \\
 Uralic        &       5 &        0.23 &     0.08 &      0.87 &      0.02 &  0.14 &      0.04 &  0.37 &      0.06 & 2.00 &     0.50 \\
 Uto-Aztecan   &       3 &        0.24 &     0.14 &      0.87 &      0.01 &  0.20 &      0.04 &  0.37 &      0.08 & 2.16 &     0.54 \\
 \hline
 TOTAL         &      70 &        0.24 &     0.12 &      0.84 &      0.05 &  0.21 &      0.06 &  0.41 &      0.10 & 2.09 &     0.70 \\
\hline
\end{tabular}}
\caption{Major results per dataset for our comparative study, comparing all 70 hand-selected language pairs in terms of Identical (phonetic strings match perfectly) and Similar word pairs (phonetic strings show SCA distance beyond our threshold of 0.5), as well as averaged SCA distances, normalized edit distance, and traditional edit distances. Highest similarities are marked in bold font, lowest similarities are shaded in gray.  }
\label{tab:comp2}
\end{table*}

When considering only the phonetic similarity scores -- which point to differences in lexeme choice when it comes to translating a concept into a given language variety -- it seems remarkable that -- apart from Indo-European, where differences make up only 6\%, we find that differences between wordlists that we would expect to represent identical language varieties show a considerable amount of variation. On average, only 83\% of all word pairs for our hand-selected sample of language pairs taken from different sources seem to be truly the same. In the remaining 17\% of cases, we find that the concepts were translated differently. Recalling that \citet{Geisler2010} report 10\% of differences with respect to lexeme choice in the Romance partition of the Indo-European dataset they compared, we can conclude that the numbers are even worse when looking at data from more language families.

\begin{figure*}[tb]
\centering
\includegraphics[width=\textwidth]{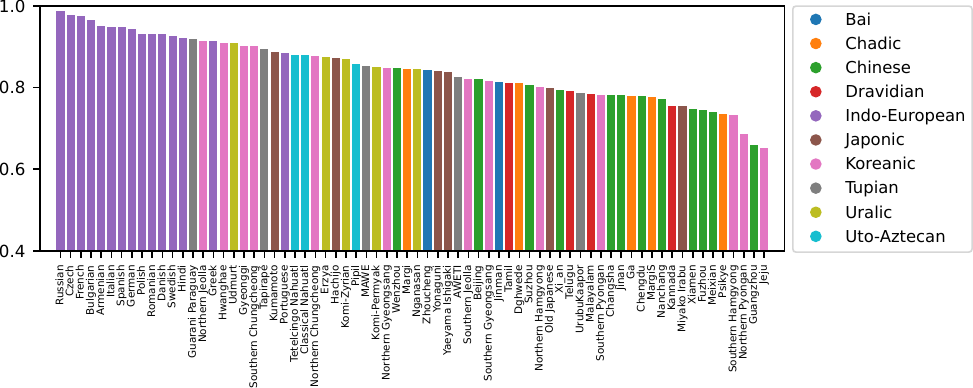}
\caption{Comparison of all language pairs in the sample.}
\label{fig:comp}
\end{figure*}

That Indo-European data shows the lowest variation in our study should not come as a surprise. First, the language family has been studied in much more detail than any other language family in the world. Second, the principles by which \citet{Heggarty2023} compiled their data have been heavily influenced by the principles laid out by the Moscow school of historical linguistics \citep{Kassian2010}, from which the second dataset on Indo-European languages in our sample was taken \citep{Starostin2005b}. Figure \ref{fig:comp} illustrates the dominance of Indo-European, in representing all 70 pair comparisons in bar charts with increasing lexical variation. As can be seen from the example, with the exception of the Indo-European datasets that show the highest similarity with respect to the translation of the concept lists into the target languages, there is no real trend that might hint to major problems with the data in particular language groups or language families. Instead, it seems that what we find in our experiments can be considered as the kind of variation that one would expect when considering the task of having independent people translate a concept list into the same language.


\subsection{Types of Variation in Concept Translation}
In order to get a better understanding regarding the sources of variation in concept translation, we carried out a more detailed analysis of the differences in the Indo-European and the Tupian data. From this comparison, we can identify two major kinds of translation differences. First, translations can point to completely different words. Second, translations may reflect the same word, but they differ morphologically. 


As an example for completely different words provided as translations for the same concept in the same target languages, consider cases like the concept \href{https://concepticon.clld.org/parameters/634}{MEAT}, translated correctly into French as \textipa{[vj\~A:d]} in the Indo-European dataset by \citet{Heggarty2023}, while the Indo-European dataset by \citet{Starostin2005b} provides two translations, \textipa{[vj\~Ad]} and \textipa{[SEK]}, the latter pointing specifically to ``human flesh''. Since one of the two forms is identical with the form in the first datasets, we count this translation difference as 0.5 in our calculations. We can see that the reason for the additional translation in the dataset by \citet{Starostin2005b} results from a lack of specification in the concept that was being compared. As an additional example, consider Tupian data for Paraguayan Guaraní, where \textipa{[gwasu]} in \citet{Galucio2015} is glossed as ``big'' (cf. \citealp[56]{Estigarribia2020}) in the original data and matched with the Concepticon gloss \href{https://concepticon.clld.org/parameters/1202}{BIG} in the Lexibank dataset. This word form may also be rendered as ``be big'' (as in \citet[224]{Gregores1967}, if one does not recognize in Paraguayan Guaraní the existence of an adjective class, for which ``little evidence exists'' \citep[15]{Estigarribia2020}. Conversely, \citet{Ferraz2021} provide the word form \textipa{[posogue]} for the concept  \href{https://concepticon.clld.org/parameters/1202}{BIG}, which could be translated to English as ``huge'' or ``gigantic'' \citep[720]{Guasch2008}. This example illustrates how subtle semantic differences influence lexical choice, leading to differences in the lexical forms selected for inclusion in a wordlist. 
 
Morphological variation also causes notable differences in concept translation. As an example, consider cases of different morphological forms in Paraguayan Guaraní. Here, some nouns have what are traditionally considered alternating roots with initial consonant \textipa{[t]}, \textipa{[h]} or \textipa{[r]}. The initial consonant may be treated as a prefix \citep[63]{Estigarribia2020}, with \textipa{[t-]} expressing, for instance, the ``absolute or non-possessed form''. \citet{Galucio2015} have a tendency to report the Paraguayan Guaraní forms with \textipa{[t-]}, while \citet{Ferraz2021} tend to only report the prefixless root. While a manual analysis would most likely treat these differences as marginal, characterizing the variants as identical translations of the original concept into the target language in question, there are cases where morphological variation can yield substantial differences in the translations. As an example, consider the two forms for \href{https://concepticon.clld.org/parameters/1424}{YELLOW} in Urubú-Ka’apór. Based on \citet[56, 116, 200, 204]{Kakumasu2007}, the representation in \citet{Galucio2015} is a suffix meaning ``yellow''  \textipa{[-ju]}, while \citet{Ferraz2021} indicate the third person form of the ``descriptive verb'' ``be yellow''  \textipa{[itawa]}.

\section{Discussion and Conclusion}\label{sec:con}
Our results confirm the findings of \citet{Geisler2010}, who emphasize that concept translation may cause larger discrepancies between wordlists compiled independently by different scholars.
While \citet{Geisler2010} report differences of about 10\% for the Romance language partition in the two Indo-European datasets they compared, our results, taking data from 9 different language families into account, show that the differences are typically even larger, with only 83\% of concept translations reflecting the same underlying word forms on average. While we have not tested the impact differences in concept translation may have on phylogenetic reconstruction, we consider our results robust enough to call for the attention of scholars who use
phylogenetic methods to answer big picture questions about human prehistory. 
 
Although we think it is far too early to discard phylogenetic methods, we think that it would be useful for phylogenetic approaches to take potential problems resulting from the concept translation stage during the wordlist compilation seriously and make sure to apply certain measures to increase the robustness of their inferences. These measures could include tests on sampling errors, as outlined in \citet{Feld2019}, involve additional tests on inter-annotator agreement \citep{McDonald2019} during concept translation, or conduct robustness tests similar to the bootstrap in phylogenetic reconstruction. 

In any case, the declared goal of phylogenetic approaches should be the same as for traditional historical linguistics. \citet{Ratcliffe2012} suggested that an ideal test of the reliability of the traditional comparative method for linguistic reconstruction would be to ``take two teams of researchers trained in the comparative method, put them in the libraries, keep them in isolation from each other and see what they come up with'' \citep[240]{Ratcliffe2012}. While it is clear that such experiments are still lacking in traditional historical linguistics, it seems important that scholars working in the field of historical language comparison, no matter if they work computationally or manually, maintain a mindset that does not take the reliability of their data for granted. 

As far as our own experiments are concerned, we are still in the early stages of this research. Additional experiments -- potentially even including additional data from language families that do not feature in the sample presented here -- will be needed to get a better understanding of the potential impact of concept translation on phylogenetic analysis. It is possible that phylogenetic methods turn out to be robust enough to yield similar results in terms of subgrouping and divergence time estimates, even if the wordlists that they employed show a certain amount of differences in the translations. Without detailed analyses, however, we cannot be sure, and should not exclude the possibility that concept translation has a direct impact on the results of phylogenetic reconstruction analyses. 
 
In addition to phylogenetic reconstruction, it would also be important to test to which degree concept translation might influence the results of other studies that make use of multilingual wordlists. Among these, the most important candidates that we can identify are global studies on sound symbolism \citep{Wichmann2010a,Johansson2020}, and studies that make use of cross-linguistic colexification data \citep{Jackson2019,Tjuka2024a}. In any case, the last word on the robustness of multilingual wordlists has not yet been spoken, and more tests are needed to understand the full implications of the findings that we reported in this study.  

\section*{Supplementary Material}
All data and code needed to replicate this study are submitted together with the study along with detailed instructions, they can be accessed via the Open Science Framework (URL: \href{https://osf.io/8sc5h/?view_only=2f3fed5b90894d28b41ea907f247efd9}{https://osf.io/8sc5h/?view\_only=2f3fed5b90894 d28b41ea907f247efd9}).
\section*{Limitations}
The most important limitation to our current study is the uneven distribution of data across language families and varieties. For instance, we have 15 comparative varieties for Indo-European, while we only have two for Bai. There is also a disparity in the number of matching concepts between matching language varieties which allows us to achieve a more extensive sample of certain paired varieties. At the moment, we do not see how these limitations could be addressed consistently. In the long run, it seems that we must try to increase the number of comparisons by trying to identify more datasets that were independently compiled for the same languages.

\section*{Acknowledgements}
This project was supported by the ERC Consolidator Grant ProduSemy (PI Johann-Mattis List,
Grant No. 101044282, see \url{https://doi.org/10.3030/101044282}). Views and opinions expressed
are, however, those of the authors only and do not
necessarily reflect those of the European Union or
the European Research Council Executive Agency
(nor any other funding agencies involved). Neither the European Union nor the granting authority
can be held responsible for them. 

 

\label{sec:bibtex}


\bibliography{custom}



\end{document}